\documentclass[10pt,twocolumn,letterpaper]{article}

\usepackage{iccv}              
\definecolor{iccvblue}{rgb}{0.21,0.49,0.74}
\usepackage{pifont}
\usepackage[pagebackref,breaklinks,colorlinks,allcolors=iccvblue]{hyperref}
\usepackage{multirow, array}
\usepackage{makecell}
\usepackage{booktabs} 
\usepackage{subcaption}
\usepackage{adjustbox}
\newcolumntype{C}[1]{>{\centering\arraybackslash}p{#1}}

\title{JWB-DH-V1: Benchmark for Joint Whole-Body Talking Avatar and Speech Generation Version I}

\author{
Xinhan Di\textsuperscript{1}\thanks{\;These authors contributed equally to this work.}\hspace{0.6em}\raisebox{-0.3ex}{\ding{41}} \quad
Kristin Qi\textsuperscript{2}\footnotemark[1] \quad
Pengqian Yu\textsuperscript{3} \\
\textsuperscript{1}Independent Researcher, China \\
\textsuperscript{2}Computer Science, University of Massachusetts Boston, USA \\
\textsuperscript{3}National University of Singapore, Singapore \\
{\tt\small deepearthgo@gmail.com, yanankristin.qi001@umb.edu, yupengqian1989@gmail.com}
}
\begin{document}
\maketitle
\thispagestyle{plain}  
\pagestyle{plain}  
\begin{abstract}
Recent advances in diffusion-based video generation have enabled photo-realistic short clips, but current methods still struggle to achieve multi-modal consistency when jointly generating whole-body motion and natural speech. Current approaches lack comprehensive evaluation frameworks that assess both visual and audio quality, and there are insufficient benchmarks for region-specific performance analysis. To address these gaps, we introduce the Joint Whole-Body Talking Avatar and Speech
Generation Version I(JWB-DH-V1), comprising a large-scale multi-modal dataset with 10,000 unique identities across 2 million video samples, and an evaluation protocol for assessing joint audio-video generation of whole-body animatable avatars. Our evaluation of SOTA models reveals consistent performance disparities between face/hand‑centric and whole‑body performance, which incidates essential areas for future research. The dataset and evaluation tools are publicly available at \href{https://github.com/deepreasonings/WholeBodyBenchmark}{https://github.com/deepreasonings/WholeBodyBenchmark}.
\end{abstract}    
\section{Introduction}
\label{sec:intro}
Diffusion-based video generation has emerged as a powerful approach for synthesizing realistic videos~\cite{singer2022make}. Researchers are actively exploring cascaded and latent-space diffusion pipelines to address existing limitations~\cite{li2023videogen,zhang2023i2vgen,wang2024lavie}. However, in the context of digital humans~\cite{prajwal2020lip,min2022styletalker}, current diffusion models still struggle to synthesize realistic full-body motion and appearance effectively~\cite{cui2024hallo2, cui2024hallo3,ji2024sonic,guan2023stylesync,meng2024echomimicv2,tu2024stableanimator}. These models rarely achieve synchronized audio and video generation for full-body speakers. This gap highlights the need for integrated systems capable of producing high-quality full-body motion synchronized naturally with speech, which is the focus of the Joint Whole-Body Talking Avatar and Speech Generation Dataset.

Moreover, creating fully natural whole-body talking avatars remains a substantial challenge. Advanced talking head models generally concentrate on facial and upper-body regions~\cite{cui2024hallo2, cui2024hallo3,ji2024sonic,guan2023stylesync} and fail to synthesize full-body motions or gestures aligned with speech~\cite{meng2024echomimicv2,tu2024stableanimator}. Furthermore, existing methods typically do not achieve stable and joint audio-video generation from a single frame~\cite{google2025veo3}. These limitations reveal the importance of establishing a benchmark specifically designed to evaluate and advance methods that integrate facial animation with whole-body motion, ensuring coherence with the generated speech audio.

\begin{figure*}[!t] 
    \centering
    \includegraphics[width=0.73\textwidth]{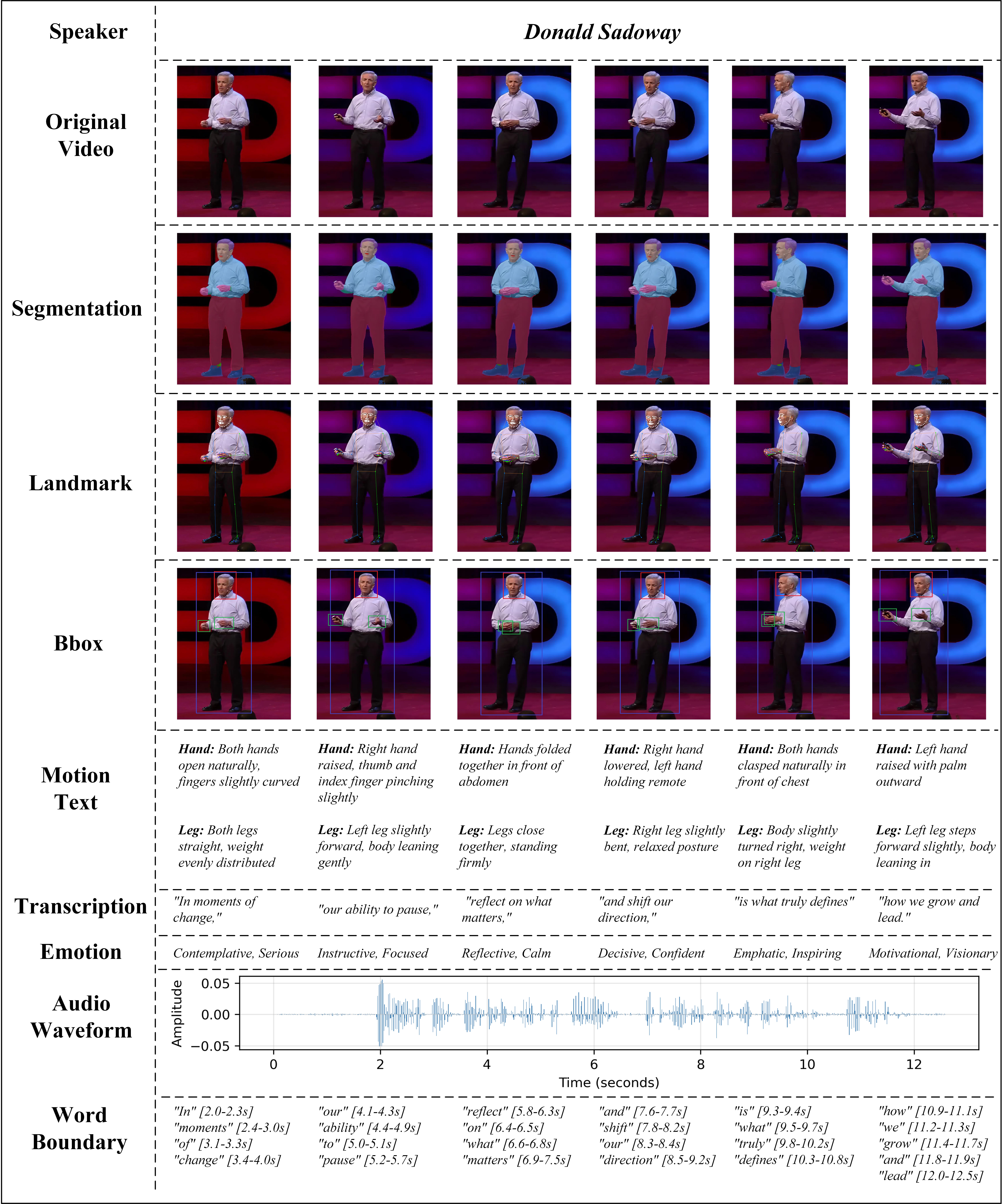}
     \caption{Illustration of our dataset annotations. Each column represents a key frame. From top to bottom: the original frame, body segmentation, landmark, bounding box annotations for key regions (hands, legs, whole body), motion text describing pose semantics, corresponding speech transcription, the motion flag of each transcription, ground truth audio and word boundary. The dataset captures fine-grained multi-modal alignment between body postures, hand gestures, leg stance, and spoken language with its audio over time.}
    \label{fig:1}
\end{figure*}

\begin{figure*}[!t] 
   \centering
\includegraphics[width=0.95\textwidth]{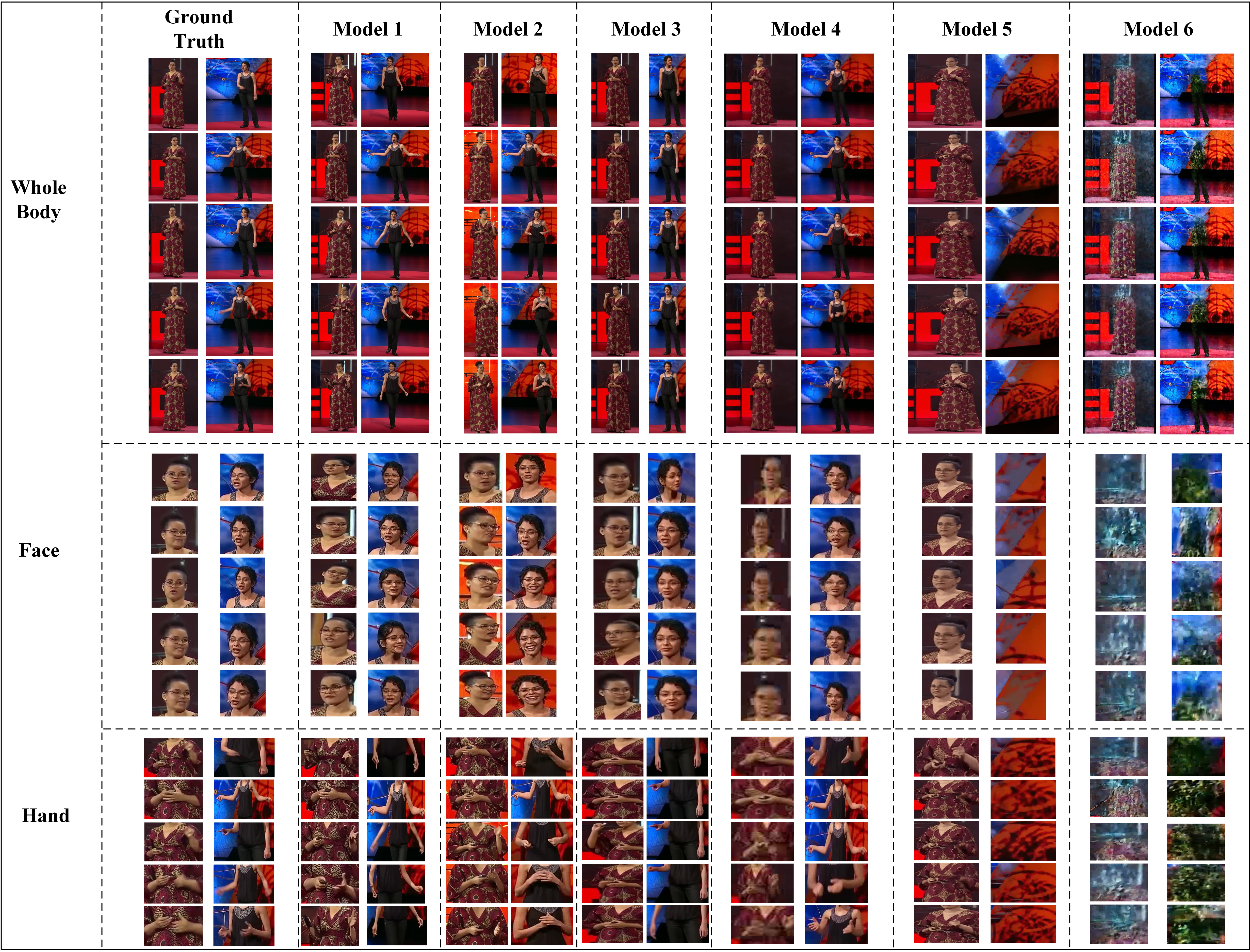}
   \caption{Visual comparison across full body, face, and hand regions. Model 1(Wan~\cite{wan2025}),2(OpenS~\cite{opensora2}),3(Hun~\cite{kong2024hunyuanvideo}), 4(ecv2/w~\cite{meng2024echomimicv2}) show good performance in structure, motion, and identity preservation. Model 5(ecv2/wo~\cite{meng2024echomimicv2}),6(Ha3/wo~\cite{cui2024hallo3}) show noticeable artifacts and degradation across all regions.}
   \label{fig:2}
\end{figure*}
\section{Related Work}
\label{sec:relatedwork}
\subsection{Diffusion-Based Video Generation and Talking Head Generation}
Diffusion-based models have been extended from images to video~\cite{ho2022video}, with Stable Video Diffusion as a foundation~\cite{blattmann2023stable}. Later works incorporate conditioning and multi-scale processing~\cite{kong2024hunyuanvideo,yang2024cogvideox,qiu2025skyreels,huang2025step}, but lack audio-visual alignment for human-centric generation. Lip-sync discriminators guide GANs or diffusion models for facial animation~\cite{prajwal2020lip,zhou2021pose,min2022styletalker}, yet these approaches are spatially constrained to head or upper body~\cite{jiang2024loopy,li2024latentsync,cui2024hallo3,ji2024sonic,guan2023stylesync}, rather than enabling full-body audio-driven generation.
\subsection{Joint Video-Audio Generation}
Recent advances video and audio synthesis include DeepMind's Veo-3 \cite{google2025veo3} and unified Diffusion Transformers \cite{liu2025javisdit,zhao2025uniform}. These systems enable synchronized multi-modal content creation. However, current joint audio-video generation approaches primarily emphasize general alignment, rather than fine-grained synchronization between speech and body movements, which is essential for digital human applications. Moreover, achieving the stable joint audio-video generation from a single input frame remains experimental. To address these challenges, we introduce the Whole-Body Joint Talking Avatar and Speech Generation benchmark that is designed to advance research in synchronized and high-quality audio-video generation.

\begin{table*}[t]
\centering
\caption{Comparison of different text-to-speech (TTS) models on a range of speech synthesis tasks.}
\scriptsize
\begin{adjustbox}{max width=\textwidth}
\begin{tabular}{lcccccccc}
\toprule
\textbf{Model} & \multicolumn{2}{c}{\textbf{Emotions}} & \multicolumn{2}{c}{\textbf{Paralinguistics}} & \multicolumn{2}{c}{\textbf{Complex Pronunciation}} & \multicolumn{2}{c}{\textbf{Syntactic Complexity}} \\
&  WER & Win-Rate & WER & Win-Rate & WER & Win-Rate & WER & Win-Rate \\
\midrule
\textit{gpt-4o-mini-tts (baseline)} & 5.76 & - & 24.59 & - & \textbf{28.92} & - & 6.09 & -  \\
\midrule
Suno Bark \cite{SunoAI_Bark_2023}  & 6.34 & 0.00\% & 29.83 & 6.67\% & 55.81 & 8.42\% & 6.02 & 15.29\% \\
Tortoise-TTS \cite{Betker2023TorToiSe}  & 15.89 & 3.71\% & 59.72 & 5.08\% & 59.84 & 5.47\% & 5.13 & 17.42\%  \\
MiniCPM \cite{Hu2024MiniCPM} & 17.62 & 8.49\% & 26.53 & 6.45\% & 61.12 & 8.11\% & 8.83 & 16.28\% \\
Qwen 2.5 Omni \cite{Xu2025Qwen25Omni} & 8.47 & 31.13\% & 23.06 & 24.02\% & 47.69 & 13.37\% & 9.01 & 28.63\% \\
\midrule
11Labs eleven multilingual v2 \cite{ElevenLabs_ElevenMultilingualV2_2023} & \textbf{4.68} & 20.38\% & 22.56 & 35.41\% & 33.49 & 14.83\% & 9.12 & 33.92\% \\
\textbf{gpt-4o-mini-audio-preview} \cite{OpenAI_GPT4oMiniAudioPreview_2024} & 5.79 & 58.31\% & 24.97 & 60.68\% & 35.58 & 37.63\% & 10.81 & 53.72\%  \\
\bottomrule
\end{tabular}
\end{adjustbox}
\label{tab:audio_table}
\end{table*}
\begin{table*}[t]
\centering
\scriptsize
\setlength{\tabcolsep}{5pt}
\renewcommand{\arraystretch}{0.9}
\caption{Evaluation of eight models across three regions and twelve metrics.
SC: Subject Consistency, BC: Background Consistency, MS: Motion Smoothness,
DD: Dynamic Degree, AQ: Aesthetic Quality, IQ: Imaging Quality,
FID/FVD: Fréchet (Video) Distance, SSIM: Structural Similarity,
PSNR: Peak Signal-to-Noise Ratio, E-FID: Enhanced FID, CSIM: Cosine Similarity.
\textbf{Bold} = best among generative models (ground-truth excluded).}
\label{tab:score_table}
\begin{tabular}{llccccccccc}
\toprule
Region & Metric & GT & Step\cite{huang2025step} & Hun\cite{kong2024hunyuanvideo} & Wan\cite{wan2025} & OpenS\cite{opensora2} & Ha3/w\cite{cui2024hallo3} & Ha3/wo\cite{cui2024hallo3} & ECV2/w\cite{meng2024echomimicv2} & ECV2/wo\cite{meng2024echomimicv2} \\
\midrule
\multirow{12}{*}{Whole Body}
 & SC   & 100 & 93.45 & 92.54 & \textbf{96.83} & 97.12 & 20.20 & 10.74 & 29.16 & 7.08 \\
 & BC   & 100 & 94.50 & 94.30 & \textbf{96.31} & 98.00 & 19.91 & 10.93 & 31.25 & 6.49 \\
 & MS   & 100 & 98.71 & 98.50 & \textbf{99.69} & 90.69 & 20.60 & 11.37 & 29.75 & 7.14 \\
 & DD   & 100 & 67.39 & \textbf{69.34} & 59.89 & 60.43 & 4.65 & 4.00 & 9.81 & 0.09 \\
 & AQ   & 100 & \textbf{48.32} & 43.17 & 46.73 & 32.88 & 10.01 & 3.84 & 14.27 & 2.52 \\
 & IQ   & 100 & 61.99 & \textbf{63.54} & 54.67 & 42.57 & 9.88 & 5.09 & 16.53 & 3.29 \\
 & FID  & 0.0 & 172.47 & 208.75 & \textbf{92.76} & 121.53 & 491.43 & 568.68 & 508.19 & 572.14 \\
 & FVD  & 0.0 & 1356.49 & 1567.94 & \textbf{750.51} & 896.94 & 2470.14 & 4366.39 & 2519.77 & 3791.25 \\
 & SSIM & 1.0 & 0.573 & 0.543 & \textbf{0.660} & 0.623 & 0.209 & 0.016 & 0.216 & 0.028 \\
 & PSNR & $\infty$ & 17.74 & 16.06 & \textbf{20.16} & 19.48 & 6.24 & 0.83 & 6.17 & 1.29 \\
 & E-FID& 0.0 & \textbf{187.58} & 216.91 & 196.39 & 218.78 & 484.08 & 634.60 & 492.17 & 712.91 \\
 & CSIM & 1.0 & 0.760 & 0.811 & \textbf{0.896} & 0.876 & 0.303 & 0.053 & 0.299 & 0.092 \\
\midrule
\multirow{12}{*}{Face}
 & SC   & 100 & 65.06 & 55.07 & \textbf{66.51} & 52.34 & 14.92 & 12.23 & 14.79 & 8.29 \\
 & BC   & 100 & \textbf{74.04} & 68.05 & 72.26 & 63.55 & 15.73 & 15.15 & 18.16 & 9.27 \\ 
 & MS   & 100 & 63.12 & 56.78 & \textbf{63.40} & 49.74 & 14.66 & 12.15 & 12.73 & 8.16 \\
 & DD   & 100 & 14.91 & 4.15 & \textbf{25.12} & 7.69 & 7.21 & 4.55 & 2.42 & 0.21 \\
 & AQ   & 100 & \textbf{28.18} & 21.26 & 25.09 & 19.76 & 5.79 & 2.99 & 5.71 & 5.69 \\
 & IQ   & 100 & \textbf{39.40} & 34.19 & 33.68 & 30.90 & 6.98 & 4.60 & 8.49 & 6.29 \\
 & FID  & 0.0 & 300.98 & 291.21 & \textbf{254.03} & 275.50 & 425.82 & 720.22 & 437.27 & 801.27 \\
 & FVD  & 0.0 & 1811.92 & 1351.71 & 1591.22 & \textbf{1221.85} & 2794.85 & 3255.19 & 2651.92 & 3047.27 \\
 & SSIM & 1.0 & 0.372 & 0.322 & \textbf{0.417} & 0.388 & 0.127 & 0.010 & 0.139 & 0.027 \\
 & PSNR & $\infty$ & 13.54 & 14.47 & \textbf{16.99} & 16.29 & 6.75 & 0.89 & 6.88 & 1.25 \\
 & E-FID& 0.0 & 339.18 & 347.24 & \textbf{273.97} & 359.85 & 451.76 & 492.67 & 439.92 & 513.58 \\
 & CSIM & 1.0 & 0.736 & 0.787 & \textbf{0.866} & 0.863 & 0.385 & 0.065 & 0.401 & 0.062 \\
\midrule
\multirow{12}{*}{Hand}
 & SC   & 100 & \textbf{66.83} & 48.15 & 51.27 & 49.16 & 13.65 & 2.13 & 14.19 & 3.58 \\
 & BC   & 100 & \textbf{77.98} & 64.98 & 57.45 & 32.45 & 14.76 & 2.56 & 18.21 & 4.27 \\
 & MS   & 100 & 46.36 & \textbf{49.65} & 42.49 & 43.30 & 13.25 & 0.49 & 9.27 & 2.79 \\
 & DD   & 100 & 48.14 & \textbf{52.85} & 41.71 & 43.89 & 13.05 & 0.91 & 9.52 & 1.84 \\
 & AQ   & 100 & \textbf{24.22} & 19.02 & 18.38 & 16.40 & 5.08 & 0.39 & 5.08 & 2.74 \\
 & IQ   & 100 & \textbf{38.40} & 27.69 & 23.84 & 21.50 & 6.51 & 0.89 & 10.24 & 1.94 \\
 & FID  & 0.0 & 327.10 & 301.94 & 331.51 & \textbf{278.95} & 574.40 & 921.16 & 588.72 & 975.92 \\
 & FVD  & 0.0 & 1597.65 & 1734.98 & 1561.58 & \textbf{1534.50} & 7750.48 & 10671.79 & 8958.27 & 9472.28 \\
 & SSIM & 1.0 & 0.316 & 0.262 & \textbf{0.367} & 0.342 & 0.098 & 0.008 & 0.106 & 0.019 \\
 & PSNR & $\infty$ & 12.98 & 11.45 & 13.09 & \textbf{15.50} & 6.13 & 0.55 & 7.27 & 0.49 \\
 & E-FID& 0.0 & 277.47 & 255.53 & \textbf{240.48} & 264.98 & 445.59 & 713.46 & 525.18 & 741.03 \\
 & CSIM & 1.0 & 0.681 & 0.737 & \textbf{0.796} & 0.794 & 0.366 & 0.041 & 0.427 & 0.152 \\
\bottomrule
\end{tabular}
\end{table*}
\section{Method}
\subsection{Dataset Curation}
We curate the dataset with \textbf{10,000 unique identities}, each appearing in approximately \textbf{200 different scene configurations}, totaling \textbf{2 million samples of video clips}. The annotation for each sample is demonstrated in Figure~\ref{fig:1}. \textbf{20,000 samples} are used for evaluation.

\subsection{Video Generation Sub Evaluation Protocol}
To evaluate perceptual quality and temporal stability of generated whole-body avatars, we adopt six reference-free video metrics: (1) Subject Consistency (SC): temporal appearance consistency via DINO features~\cite{caron2021emerging}, (2) Background Consistency (BC): background stability via CLIP features~\cite{radford2021learning}, (3) Motion Smoothness (MS): optical flow continuity~\cite{li2023amt}, (4) Dynamic Degree (DD): motion richness from flow magnitude~\cite{teed2020raft}, (5) Aesthetic Quality (AQ): LAION aesthetic score~\cite{laion_aesthetic_predictor}, and (6) Imaging Quality (IQ): visual clarity via MUSIQ~\cite{ke2021musiq}. These metrics provide a comprehensive evaluation of realism and coherence.
\vspace{-1pt}
\subsection{Co-Speech Sub Evaluation Protocol}
To complement the region-level evaluation of the whole-body generated avatars, we adopt six standard metrics to assess frame quality, temporal coherence, and identity preservation~\cite{cui2024hallo2, cui2024hallo3,ji2024sonic,guan2023stylesync}. These include FID~\cite{qin2025versatile}, FVD~\cite{unterthiner2018towards}, SSIM~\cite{wang2004image}, PSNR~\cite{huynh2012accuracy}, E-FID~\cite{qin2025versatile}, and CSIM~\cite{qin2025versatile}. These metrics capture fidelity, temporal stability, and identity consistency for comprehensive evaluation.
\vspace{-4pt}
\subsection{Speech Audio Sub Evaluation Protocol}
To evaluate the quality of generated audio, we use Large-Audio-Language Models (LALMs) as outlined in recent evaluation methods \cite{Manku2025EmergentTTS-Eval}. It targets aspects of speech synthesis such as prosody, pausing, expressiveness, and pronunciation that are not adequately captured by traditional metrics like Word Error Rate (WER). Therefore, alongside WER, we select Gemini 2.5 Pro \cite{ET_GoogleGemini2.5Pro_2025Jun17} as our primary LALM-based evaluator \cite{Manku2025EmergentTTS-Eval}. To evaluate a candidate text-to-speech (TTS) system $T_{i}$, we compare it against a strong reference system $T_{j}$ that has low WER to ensure high-fidelity synthesis of the benchmark utterances. We adopt a win-rate-based metric to summarize performance. Let $W(T_{i})$ denote the win-rate of system $T_{i}$ relative to the baseline $T_{j}$. $W(T_i)$ is computed as: $W(T_i) = \frac{\sum (\text{winner} = \text{index}_i) + 0.5 \cdot \sum (\text{winner} = 0)}{n}$, where index $x_{i} \in \{1, 2\}$ corresponds to the randomized label assigned to $T_{i}$, and $n$ denotes the total number of comparisons.

\section{Initial Evaluation}
We evaluate four groups of state-of-the-art (SOTA) models: (1) open-source video generation~\cite{kong2024hunyuanvideo, huang2025step, wan2025, opensora2}, (2) open-source talking avatars~\cite{meng2024echomimicv2, cui2024hallo3}, (3) open-source TTS systems~\cite{Xu2025Qwen25Omni,Hu2024MiniCPM,Betker2023TorToiSe,SunoAI_Bark_2023}, and (4) closed-source TTS systems~\cite{OpenAI_GPT4oMiniAudioPreview_2024,ElevenLabs_ElevenMultilingualV2_2023}. Table~\ref{tab:audio_table} and ~\ref{tab:score_table} report audio and video metrics performance, respectively. Figure~\ref{fig:2} shows region-specific results for the face, hands, and full body. The joint audio-video model \cite{google2025veo3} is excluded due to instability in generating synchronized outputs from a single frame.

Video generation models take an initial video frame and a text prompt to guide motions (Figure~\ref{fig:1}). Talking avatar models are driven by audio with or without pose guidance, which are referred to as Ha3/w and Ha3/wo~\cite{cui2024hallo3}. To evaluate regional generative quality, we evaluate face, hands, and whole body independently. Additionally, utterance texts with prompting strategies (e.g., emotion prompts in Figure~\ref{fig:1}) are used to evaluate each TTS system.
\section{Conclusion}
We present JWB-DH-V1 (Joint Whole-Body Talking Avatar and Speech Generation Benchmark Version 1), a large-scale public dataset with evaluation protocols for evaluating joint whole-body avatar generation. Future releases will support 60-second clips with refined audio-video annotations and will further evaluate SOTA stable joint generation models from a single initial frame.

{
    \small
    \sloppy
     \bibliographystyle{abbrvnat}
    \bibliography{main2}
}

\end{document}